\pdfoutput=1

\documentclass[11pt]{article}

\usepackage[final]{acl}
\usepackage{arydshln}
\usepackage{times}
\usepackage{latexsym}
\usepackage{hyperref}

\usepackage{tcolorbox} 

\usepackage[T1]{fontenc}

\usepackage[utf8]{inputenc}

\usepackage{microtype}
\usepackage{arydshln} 

\usepackage{inconsolata}

\usepackage{graphicx}
\usepackage{array}
\usepackage{multirow}
\usepackage{booktabs}
\usepackage{colortbl}
\usepackage{xcolor}
\usepackage{vcell}
\usepackage{arydshln}
\usepackage{amsmath}
\usepackage{amsfonts}
\usepackage{tabularx}
\usepackage{makecell}
\usepackage[normalem]{ulem}
\usepackage{CJKutf8}
\usepackage{pifont}
\usepackage{arydshln}
\usepackage{tikz}
\usepackage{pgfplots}
\usepackage{inconsolata}

\usepackage{xcolor} 
\usepackage{algorithm}
\usepackage{algpseudocode}
\newfloat{method}{htbp}{lop}
\floatname{method}{Method}

\usepackage{todonotes}
\usepackage[capitalise,noabbrev]{cleveref}
\usepackage{listings}
\usepackage{fancyvrb} 
\makeatletter
\def\adl@drawiv#1#2#3{%
        \hskip.5\tabcolsep
        \xleaders#3{#2.5\@tempdimb #1{1}#2.5\@tempdimb}%
                #2\z@ plus1fil minus1fil\relax
        \hskip.5\tabcolsep}
     
\definecolor{verbgray}{gray}{0.9}

\usepackage{colortbl}
\definecolor{lightgray}{rgb}{0.7,0.7,0.7}

\newcommand{\cdashlinelr}[1]{%
  \noalign{\vskip 2pt}   
  \cdashline{#1}[.4pt/2pt] 
  \noalign{\vskip 2pt}   
}

\definecolor{light-orange}{HTML}{fee9d4}
\definecolor{light-green}{HTML}{d8f0d3}
\definecolor{light-blue}{HTML}{dae8f5}
\definecolor{light-red}{HTML}{FBC7C4}
\definecolor{set10-red}{HTML}{e41a1c}
\definecolor{set10-blue}{HTML}{377eb8}
\definecolor{set10-green}{HTML}{4daf4a}
\definecolor{bblue}{HTML}{4F81BD}
\definecolor{rred}{HTML}{c4260b}
\definecolor{ggreen}{HTML}{098c1f}
\definecolor{ppurple}{HTML}{9F4C7C}
\definecolor{oorange}{HTML}{F79646}

\usepackage{enumitem}
\setlist[itemize,enumerate]{leftmargin=*}
\usepackage[normalem]{ulem}
\usepackage{pgfplots}
\usetikzlibrary{pgfplots.groupplots}
\pgfplotsset{compat=1.3}
\usepackage{tikz}
\usetikzlibrary{patterns}
\usepackage{xcolor}
\usepackage{todonotes}
\usepackage{listings}
\usepackage{multirow}
\usepackage{graphicx}
\definecolor{CustomBlue}{RGB}{57,83,191}

\newtcbox{\clustertab}[1]{on line, box align=base, colback={#1},colframe={#1},size=fbox,arc=2pt,top=-1.5pt, bottom=-1.5pt, left=-1.5pt, right=-1.5pt, boxrule=0pt, enlarge left by=1pt}

\newtheorem{theorem}{Theorem}[section]
\title{\includegraphics[height=1.5ex]{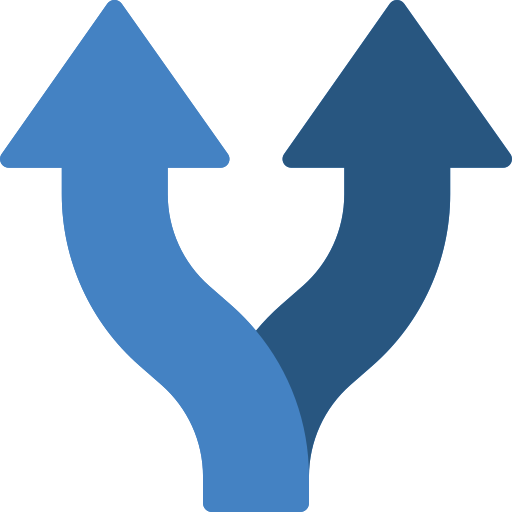} CP-Router: An Uncertainty-Aware Router Between LLM and LRM}

\makeatletter
\def\thanks#1{\protected@xdef\@thanks{\@thanks
        \protect\footnotetext{#1}}}
\makeatother

\author{
    Jiayuan Su$^{1*}$ \quad 
    Fulin Lin$^{1*}$ \quad 
    Zhaopeng Feng$^{1*}$\quad
    Han Zheng$^{1}$\quad 
    Teng Wang$^{2}$\quad\\
    \bf Zhenyu Xiao$^{3}$\quad
    \bf Xinlong Zhao$^{4}$\quad 
    \bf Zuozhu Liu$^{1{\dag}}$\quad
    \bf Lu Cheng$^{5}$\quad
    \bf Hongwei Wang$^{1{\dag}}$\quad \\
    $^{1}$Zhejiang University \quad
    $^{2}$University of Hong Kong \quad
    $^{3}$Tsinghua University \quad \\
    $^{4}$Peking University \quad
    $^{5}$University of Illinois Chicago \quad
    \\
    \texttt{\{jiayuan.23, fulin1.24, zhaopeng.23\}@intl.zju.edu.cn} \\
    \texttt{\{hongweiwang, zuozhuliu\}@intl.zju.edu.cn}
}

\thanks{$^{*}$ These authors contributed equally to this work.} 
\thanks{$^{\dag}$ \space Corresponding author.}

\begin{document}
\maketitle

\begin{abstract}

Recent advances in Large Reasoning Models (LRMs) have significantly improved long-chain reasoning capabilities over Large Language Models (LLMs). However, LRMs often produce unnecessarily lengthy outputs even for simple queries, leading to inefficiencies or even accuracy degradation compared to LLMs. To overcome this, we propose CP-Router, a training-free and model-agnostic routing framework that dynamically selects between an LLM and an LRM, demonstrated with multiple-choice question answering (MCQA) prompts. The routing decision is guided by the prediction uncertainty estimates derived via  Conformal Prediction (CP), which provides rigorous coverage guarantees. To further refine the uncertainty differentiation across inputs, we introduce Full and Binary Entropy (FBE), a novel entropy-based criterion that adaptively selects the appropriate CP threshold. Experiments across diverse MCQA benchmarks—including mathematics, logical reasoning, and Chinese chemistry—demonstrate that CP-Router efficiently reduces token usage while maintaining or even improving accuracy compared to using LRM alone. We also extend CP-Router to diverse model pairings and open-ended QA, where it continues to demonstrate strong performance, validating its generality and robustness. 
\end{abstract}

\section{Introduction}

Recent advancements, such as OpenAI's o1/o3~\citep{jaech2024openai} and DeepSeek-R1~\citep{guo2025deepseek}, have markedly improved the long-chain reasoning capabilities of large reasoning models (LRMs). While these models often outperform large language models (LLMs) on complex tasks by leveraging increased computational resources, they frequently generate unnecessarily verbose and redundant reasoning for straightforward prompts—i.e., inputs that are direct and require no multi-step inference. This inefficiency, commonly referred to as overthinking~\citep{sui2025stop, chen2025think23overthinkingo1like, ballon2025relationship}, has become a growing concern. 

\begin{figure}[t]
\centering
    \includegraphics[width=\linewidth]{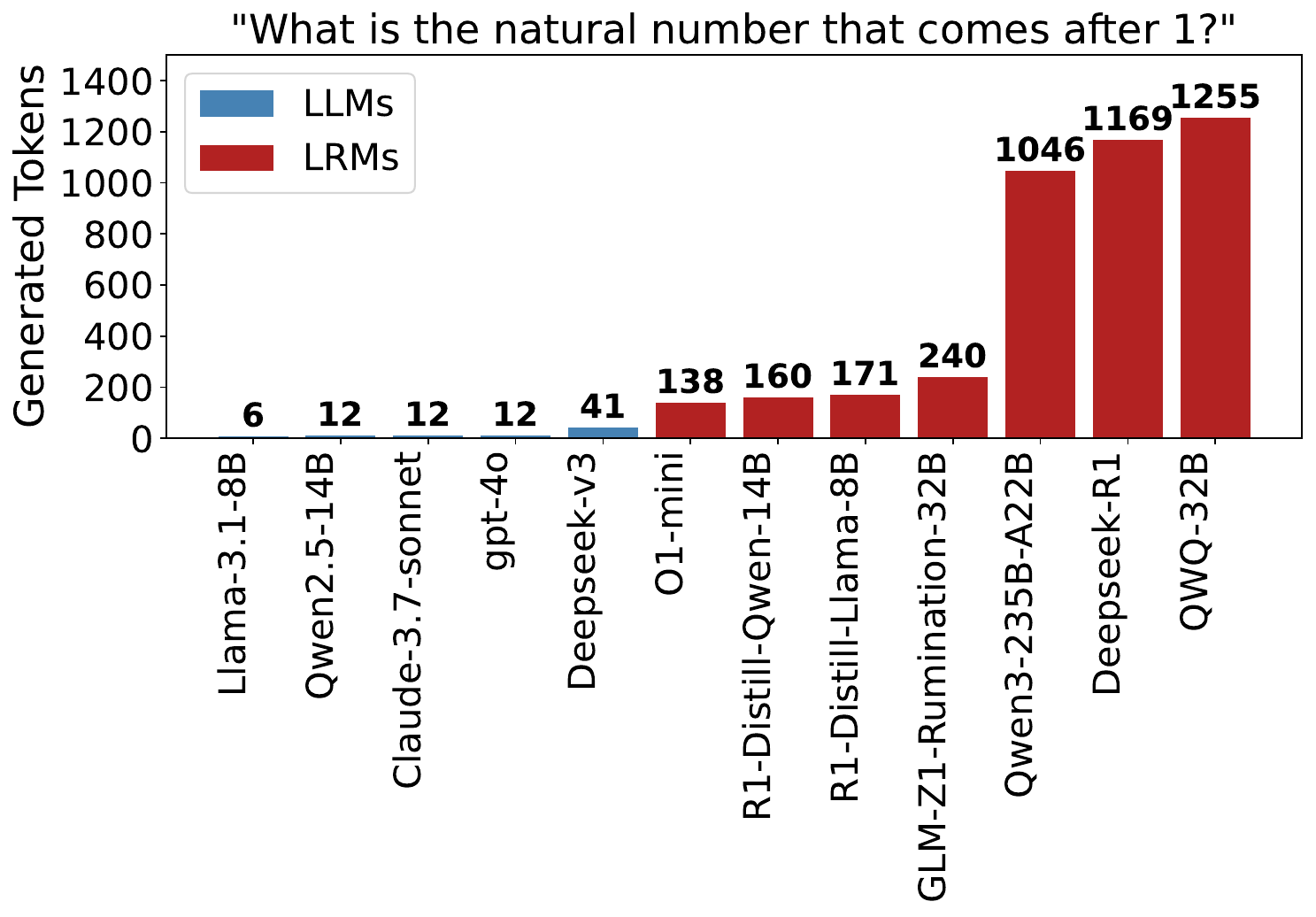}
    
\caption{Token consumption for the question "What is the natural number that comes after 1?" LLMs provide correct answers with concise token usage, whereas LRMs consume significantly more tokens, suggesting a potential "overthinking" issue.}
    \label{FIG:tokens}
    \vspace{-4mm}
\end{figure}

Recent studies~\citep{su2025between, wu2025more} indicate that excessive reasoning not only incurs additional computational overhead but can also harm accuracy. In contrast, LLMs can produce concise and often more accurate responses with significantly lower cost (as illustrated in Figure~\ref{FIG:tokens} and Figure~\ref{FIG:apss}a). This disparity reveals a critical inefficiency in current LRM behavior and raises the central question of our work:
\textit{How can we dynamically route prompts between an LLM and an LRM to minimize computational cost without sacrificing—and potentially even improving—accuracy?}

The exploration of effective LLM routing strategies remains limited. \citet{pan-etal-2024-dynathink} propose a majority-voting-based method to assess response consistency, routing inconsistent ones to a more deliberate reasoning path. However, it incurs high computational costs due to repeated sampling. More recently, advances such as Claude-3.7-Sonnet \citep{anthropic_37_2025} and Qwen3 \citep{qwen_3_2025} empower users with manual controls to switch between reasoning and standard modes. However, this places a burden on users to make effective routing decisions. These limitations highlight the urgent need for a practical and generalizable routing mechanism. An effective router is expected to exhibit the following key features: (1) \textbf{self-adaptive} — it should automatically determine whether a prompt is better suited for an LLM or an LRM, and adapt its decisions across diverse datasets; (2) \textbf{model-agnostic} — it should be compatible with model pairings comprising different families of LLMs and LRMs. (3) \textbf{lightweight} — it should require no dedicated training while maintaining low token overhead and high inference-time accuracy.

To design an effective routing strategy, we adopt Conformal Prediction (CP)~\citep{vovk2005algorithmic, angelopoulos2021gentle}, a statistical framework that quantifies the prediction uncertainty of a model for a given input. CP is training-free and model-agnostic, naturally supporting the key requirements of a router. Crucially, CP also provides strong, finite-sample theoretical guarantees: given a user-specified error rate $\alpha$, it constructs a \textit{prediction set} that contains the true label with probability at least $1 - \alpha$~\citep{vovk2005algorithmic}. This coverage guarantee holds under the exchangeability assumption and distinguishes CP from other uncertainty quantification methods that may lack such formal correctness. Moreover, the size of the calibrated prediction set serves as a direct and interpretable proxy for uncertainty—larger sets naturally correspond to higher uncertainty. These properties make CP not just a practical option, but a principled and theoretically grounded choice for prompt routing.

\begin{figure*}[t]
\centering
    \includegraphics[width=1.01\linewidth]{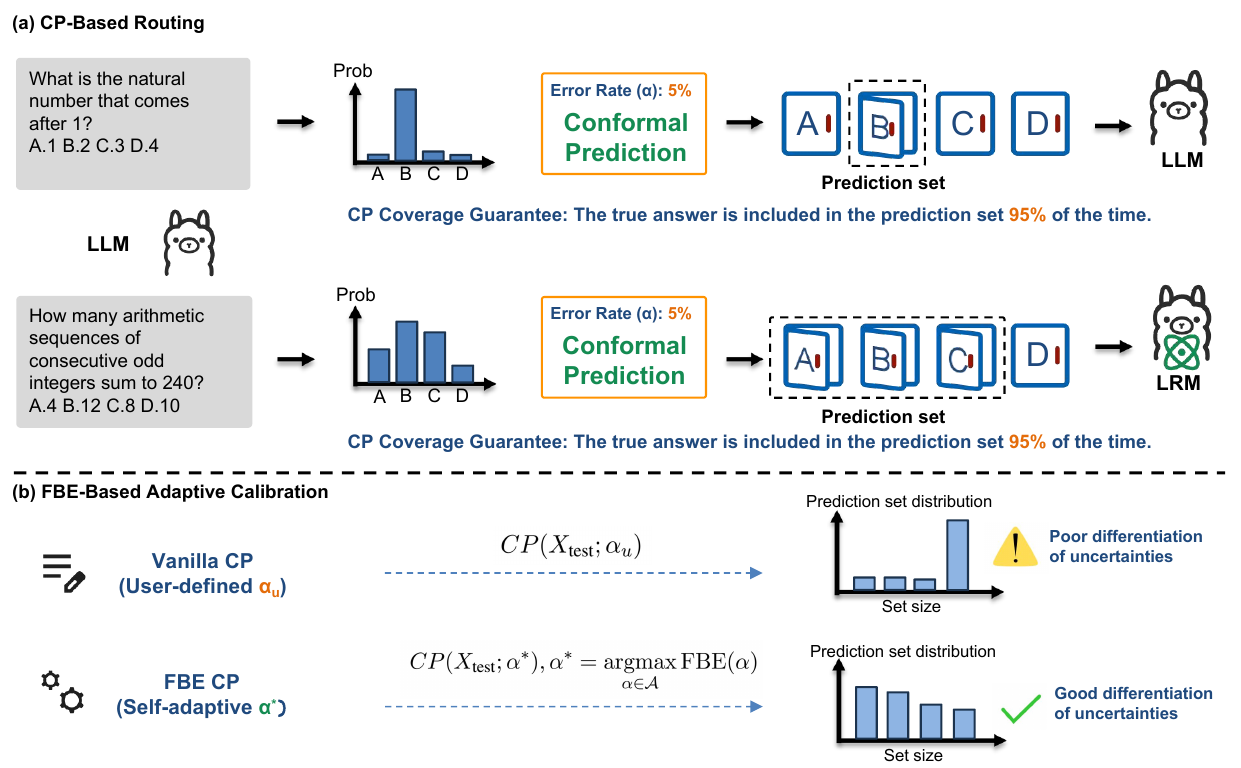}
    
\caption{Key Components of the CP-Router Framework. (a) CP-Based Routing. For each prompt, it applies CP with a target error rate $\alpha$ to generate a prediction set based on LLM output probabilities. Prompts with small prediction sets  are routed to an LLM, while those with large sets are routed to an LRM. This enables uncertainty-aware, dynamic routing. (b) FBE-Based Adaptive Calibration. Instead of using a fixed error rate, CP-Router leverages FBE to automatically select the optimal $\alpha$ that maximizes uncertainty separability, enabling more effective differentiation between easy and hard prompts—crucial for adaptive routing decisions.}
    \label{main}
    \vspace{-3mm}
\end{figure*}

We thus use CP as the backbone of our routing strategy. For each input, we obtain a prediction set from an LLM and use its size as a proxy for uncertainty. Prompts with small output sets (i.e., low uncertainty) are processed directly using the prediction from the LLM, while others are routed to the LRM. This routing strategy is illustrated in Figure~\ref{main}a. Our empirical results in Figure~\ref{FIG:apss} further support this strategy: low-uncertainty prompts are typically better handled by LLMs, while high-uncertainty prompts benefit more from LRMs.

One major challenge of applying CP in our task is selecting an appropriate error rate $\alpha$ for CP \cite{angelopoulos2021gentle}. Therefore, we propose a novel automatic calibration method that leverages a novel Full and Binary Entropy (FBE) metric to dynamically adjust $\alpha$. FBE integrates both the global and binary entropy of the prediction set size distribution, enabling a more adaptive routing strategy, as illustrated in Figure~\ref{main}b. Experiments across diverse benchmarks demonstrate that CP-Router achieves favorable accuracy and efficiency: it effectively reduces token overhead while maintaining or even improving accuracy compared to using the LRM alone. We further extend our framework to diverse model pairings and the open-ended QA task, validating the generality of CP-Router. Our main contributions are as follows:
\begin{itemize}[topsep=1pt, before=\setlength{\itemsep}{1pt}, parsep=1pt, leftmargin=*]

\item We introduce CP-Router as a novel uncertainty-aware routing mechanism that dynamically route prompts between LLM and LRM based on estimated prediction uncertainty. CP offers strong coverage guarantees and produces interpretable  prediction sets, making it particularly well-suited for prompt routing without requiring any additional training or model-specific tuning.

\item We propose a novel automatic calibration method that leverages the FBE metric to adaptively select an appropriate CP error rate, ensuring well-separated prediction uncertainties and further balancing accuracy and token efficiency.

\item Experimental results demonstrate that CP-Router reduces token consumption compared to routing all prompts to LRMs, while maintaining comparable accuracy—and even surpassing it in some cases. We validate the effectiveness of FBE and show that CP-Router generalizes well across a broad range of models and tasks.

\end{itemize}

\section{Conformal Prediction}
\label{CP}
CP \cite{vovk2005algorithmic, angelopoulos2021gentle} is a statistical framework to estimate/quantify uncertainty. Given a user-specified error rate \(\alpha\) (e.g., 10\%), CP constructs a \textit{prediction set} for each input in a principled manner (detailed in Section~\ref{cp steps}), ensuring that the true label is included with probability approximately \(1 - \alpha\) (e.g., 90\%). This guarantee is formally established in Theorem~\ref{theorem21}. Moreover, CP is applicable regardless of the underlying data distribution or model, making it particularly well-suited for our router design.

\subsection{Terminologies}

\paragraph{Score Function}
In CP, a score function (or nonconformity measure) $ S(x, y) \in \mathbb{R} $ is defined to quantify the agreement (or ``conformity'') between an input \( x \) and a candidate output \( y \). In our work, we focus on the multiple-choice question answering (MCQA) setting, and define $S(x, y)$ as \( 1 - f(y) \)~\citep{sadinle2019least}, where \( f(y) \) denotes the probability assigned to option \( y \) after applying the softmax over the logits of the candidate choices (e.g., ‘A’, ‘B’, ‘C’, and ‘D’), conditioned on the prompt. Intuitively, higher scores indicate \( y \) is more plausible given \( x \).

\begin{figure}[h]
\centering
    \includegraphics[width=0.9\linewidth]{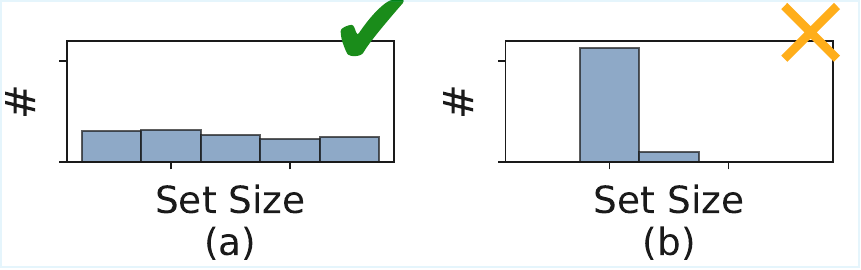}
    
\caption{Different prediction set size distributions. A manually chosen error rate $\alpha$ might lead to a poor spread, as in (b), whereas (a) illustrates a desirable distribution.}
    \label{FIG:setsize}
    \vspace{-3mm}
\end{figure}

\paragraph{Prediction Set}
\label{cp_set_size_explanation}
CP outputs a \textit{prediction set}, which contains all candidate labels deemed plausible for a given input under a user-defined error rate. The prediction set for an input \( x \) is defined as
\begin{equation}
\small
\label{eq:prediction-set}
C(x) = \{ y \in \mathcal{Y} : S(x, y) \leq \hat{q} \},
\end{equation}
where \( S(x, y) \) is the score function and \( \hat{q} \) is the quantile threshold computed from the calibration set. The size of the prediction set can be interpreted in two complementary ways. First, under the same error rate, a larger prediction set indicates higher uncertainty for the input. Second, the distribution of prediction set sizes across different inputs reflects the adaptivity of the CP procedure, as shown in Figure~\ref{FIG:setsize}. A wider spread of set sizes is desirable, since it means that the
procedure is effectively distinguishing between easy and hard inputs\cite{angelopoulos2021gentle}.

\subsection{Formal Definition} 
Let $(X, Y)$ be a sample, where $X$ represents features and $Y$ represents the outcome. Given a calibration set and a test set drawn independently and identically from the same underlying distribution, we denote the calibration set as $(X_i, Y_i)_{i=1,\ldots,n}$ and the test set as $(X_{\text{test}}, Y_{\text{test}})$. CP presents the following nesting property:
\begin{equation}
\small
\label{nesting-pro}
\alpha_1 > \alpha_2 \Rightarrow C_{1-\alpha_1}(X) \subseteq C_{1-\alpha_2}(X).
\end{equation}

\begin{theorem} [Conformal Coverage Guarantee]
\label{theorem21}
Suppose $(X_i, Y_i)_{i=1,...,n}$ and $(X_{\text{test}}, Y_{\text{test}})$ are independent and identically distributed (i.i.d.). $C: \mathcal{X} \rightarrow 2^{\mathcal{Y}}
$ is a set-valued mapping satisfying the nesting property in Eq. \ref{nesting-pro}. The following holds: 
\begin{equation}
\small
\label{marginal}
\mathbb{P}\left(Y_{\text{test}} \in \mathcal{C}(X_{\text{test}})\right) \geq 1 - \alpha
\end{equation}
where \(\alpha \in (0, 1)\) is the user-defined error rate, and $\mathcal{C}(X_{\text{test}})$ is the prediction set for input $X_{\text{test}}$.
\end{theorem}

\subsection{CP Steps}
\label{cp steps}
The standard procedure involves four steps:
\begin{itemize}[topsep=1pt, before=\setlength{\itemsep}{1pt}, parsep=1pt, leftmargin=*]
\item Establish heuristic uncertainty notions, e.g., token logits/probabilities in LLM.
\item  Define the score function $ S(x, y) \in \mathbb{R} $. We define the score as \( 1 - f(y) \), where \( f(y) \) denotes the probability assigned to the option $y$.
\label{step3}
\item Compute $ \hat{q} $ as the $  \frac{\lceil(n+1)(1-\alpha) \rceil}{n} $ quantile of the scores, where $n$ is the number of calibration samples, and $\alpha$ is a user-defined error rate. 
\label{step4}
\item Use $ \hat{q} $ to generate prediction sets for new samples: $ C(X_{\text{test}}) = \{Y : S(X_{\text{test}}, Y) \leq \hat{q}\} $.  These prediction sets may vary in size depending on the uncertainty of the predictions.
\end{itemize}

Finally, we obtain statistically grounded prediction sets that quantify prediction uncertainty and guide our \textbf{CP-Router}, which uses their size to route between LLMs and LRMs.

\section{CP-Router}
We first verify that the prediction set produced by CP serves as a reliable proxy for uncertainty. Then, we employ an FBE-guided calibration strategy to adaptively tune the error rate, enhancing uncertainty separation across inputs. The calibrated error rate is used to perform CP, constructing prediction sets that enable effective routing and achieve a strong balance between accuracy and efficiency.

\subsection{Uncertainty Estimation via CP}
\label{vanillacp}
\begin{figure}[t]
\centering
    \includegraphics[width=\linewidth]{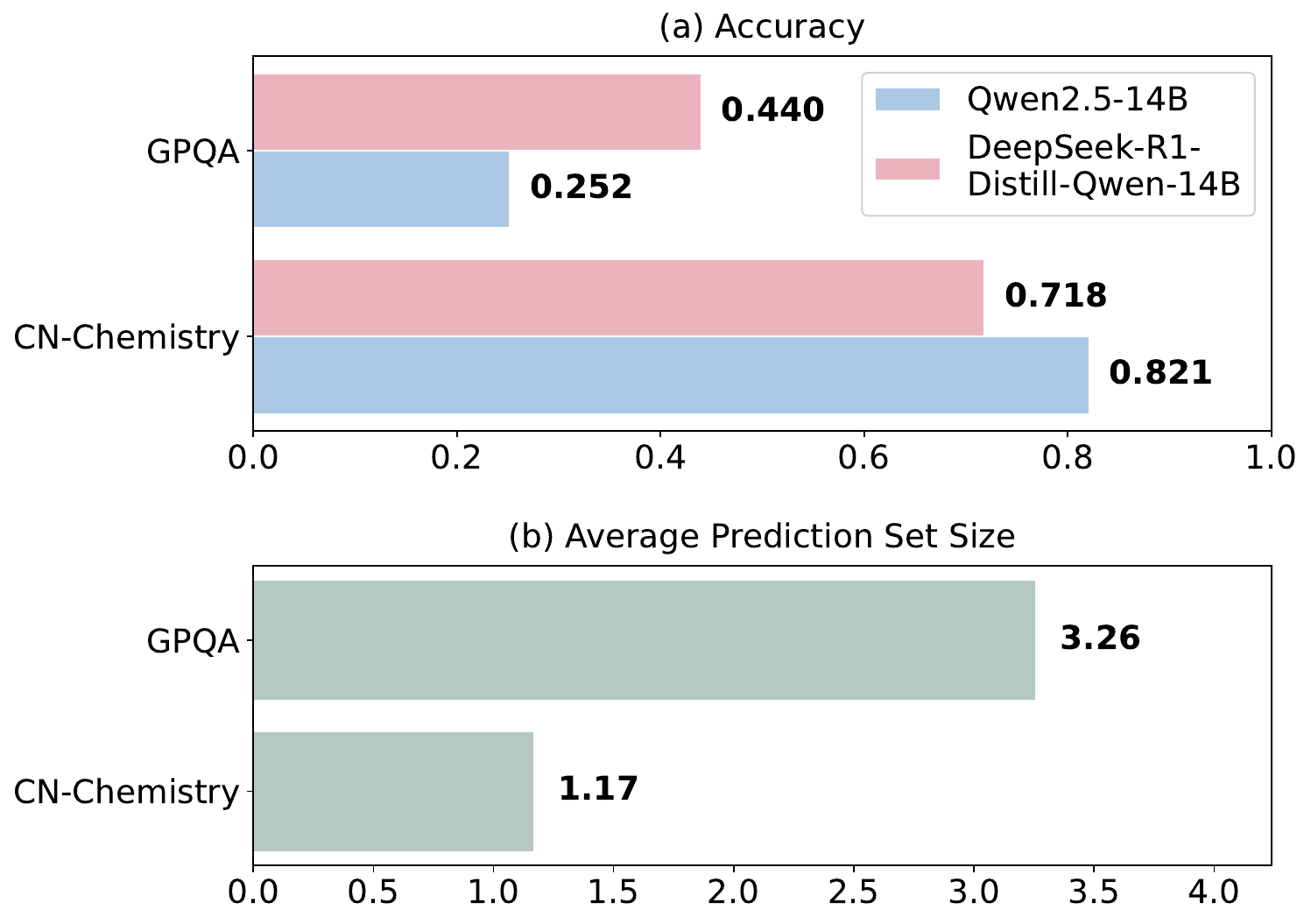}
    
\caption{(a) Accuracy comparison between Qwen2.5-14B (LLM) and DeepSeek-R1-Distill-Qwen-14B (LRM) on the GPQA and CN-Chemistry benchmarks. The LRM performs better on GPQA, while the LLM achieves higher accuracy on CN-Chemistry. (b) Average Prediction Set Size (APSS) of Qwen2.5-14B under an error rate of 0.2 on both datasets. GPQA exhibits a larger APSS, suggesting higher prediction uncertainty.}

    \label{FIG:apss}
    \vspace{-2mm}
\end{figure}

Building on the CP framework discussed in Section~\ref{CP}, we estimate prediction uncertainty for each input under the LLM to determine whether a question should be routed to the LLM or the LRM. The key intuition is that questions with low prediction uncertainty can be confidently handled by the LLM, while those with high uncertainty should be routed to the LRM for further consideration. To validate this intuition, we conduct experiments using Qwen2.5-14B~\citep{qwen2025qwen25technicalreport} (LLM) and DeepSeek-R1-Distill-Qwen-14B~\citep{guo2025deepseek} (LRM) on two benchmarks: GPQA~\cite{rein2024gpqa} and CN-Chemistry~\cite{zhang2024chemllm}. We first evaluate the accuracy of both models on these tasks, and then apply CP under Qwen2.5-14B to compute the Average Prediction Set Size (APSS). As illustrated in Figure~\ref{FIG:apss}, Qwen2.5-14B achieves higher accuracy on CN-Chemistry, which exhibits a smaller APSS—suggesting low prediction uncertainty. In contrast, DeepSeek-R1-Distill-Qwen-14B performs better on GPQA, where the APSS is larger—indicating higher uncertainty. These results empirically support our intuition.

Since our framework is designed for MCQA prompts, we extract the logits corresponding to the answer choices (e.g., ‘A’, ‘B’, ‘C’, and ‘D’) for each prompt and apply the softmax function to obtain a probability distribution over the options. These probabilities define the nonconformity scores used in CP procedure. CP then constructs a prediction set that includes the plausible answer options under a given error rate $\alpha$. We adopt a prediction set size-aware routing rule: if the prediction set is small (e.g., of size 1), this indicates low uncertainty, and the prompt is handled by the LLM. Otherwise, it is routed to the LRM for more deliberate processing. 

\subsection{Adaptive CP Routing}
However, a central challenge in using CP for routing lies in selecting the error rate $\alpha$~\citep{angelopoulos2021gentle}, which determines the target coverage of the prediction sets. Although $\alpha$ is user-defined, there are no established principles for tuning it specifically for routing purposes. As illustrated in Section~\ref{CP}, the choice of $\alpha$  influences the prediction set sizes and, in turn, affects routing decisions. An inappropriate $\alpha$ can lead to suboptimal adaptivity: a small $\alpha$ tends to produce large APSS, causing the LLM to reject many prompts and overburden the LRM; conversely, a large $\alpha$ may yield overly small APSS, reducing the engagement of the LRM even when necessary. Additionally, certain $\alpha$ values can result in prediction set distributions with poor separation between high- and low-uncertainty prompts~\citep{angelopoulos2021gentle}, as shown in Figure~\ref{FIG:setsize}b, limiting the flexibility of the router. In our experiments, we explore the impact of different error rates in our experiments (Table~\ref{ablations}), and observe that the choice of $\alpha$ plays a critical role in balancing accuracy and token efficiency.

\begin{table*}[t]
\centering
\small
\setlength{\tabcolsep}{1.4mm}
\renewcommand{\arraystretch}{1.2}
\resizebox{\textwidth}{!}{%
\begin{tabular}{@{}lc ccc ccc ccc ccc ccc ccc @{}}
\toprule
\multirow{2}{*}{\textbf{Method}} & \multirow{2}{*}{\textbf{Threshold}} 
& \multicolumn{3}{c}{\textbf{Elementary Math}} 
& \multicolumn{3}{c}{\textbf{High School Math}} 
& \multicolumn{3}{c}{\textbf{College Math}}
& \multicolumn{3}{c}{\textbf{LogiQA}} 
& \multicolumn{3}{c}{\textbf{GPQA}} 
& \multicolumn{3}{c}{\textbf{STEM-MCQA}} \\
\cmidrule(lr){3-5} \cmidrule(lr){6-8} \cmidrule(lr){9-11} \cmidrule(lr){12-14} \cmidrule(lr){15-17} \cmidrule(lr){18-20} 
& & Acc & TRR & $U_{\text{token}}$ 
  & Acc & TRR & $U_{\text{token}}$ 
  & Acc & TRR & $U_{\text{token}}$ 
  & Acc & TRR & $U_{\text{token}}$
  & Acc & TRR & $U_{\text{token}}$ 
  & Acc & TRR & $U_{\text{token}}$ \\
\midrule

\multicolumn{20}{c}{\textbf{Llama-3.1-8B} and \textbf{DeepSeek-R1-Distill-Llama-8B}} \\
LLM
& - & 41.6 & - & - & 31.9 & - & - &  31.2 & - & - & 35.9 & - & - & 29.6 & - & - & 24.4 & - & - \\
LRM
& - & 79.9 & - & 38.3 & 70.4 & - & 38.5 &  58.8 & - & 27.6 & 46.4 & - & 10.5 & 34.0 & - & 4.4 & 65.0 & - & 40.6 \\
\cdashline{1-20}
\multirow{4}{*}{Random}
& 0.2 & 72.3 & 19.8 & 38.3 & 63.0 & 19.3 & 38.5 & 53.8 & 16.0 & 26.9 & 44.7 & 19.7 & 11.0 & 33.1 & 19.9 & 4.4 & 57.3 & 19.4 & 40.8 \\
& 0.3 & 69.0 & 29.4 & 38.8 & 59.1 & 29.4 & 38.5 & 51.5 & 25.3 & 27.2 & 43.5 & 29.1 & 10.8 & 32.7 & 29.4 & 4.4 & 53.4 & 28.9 & 40.8 \\
& 0.4 & 65.7 & 39.3 & 39.8 & 55.6 & 39.1 & 38.8 & 49.3 & 34.5 & 27.6 & 42.8 & 39.3 & 11.4 & 33.1 & 38.7 & 5.7 & 49.4 & 38.9 & 40.9 \\
& 0.5 & 61.9 & 49.4 & 40.1 & 51.5 & 49.0 & 38.4 & 45.5 & 45.0 & 26.0 & 41.8 & 48.9 & 11.5 & 32.1 & 48.4 & 4.8 & 45.9 & 42.5 & 37.4 \\
\cdashline{1-20}
\multirow{3}{*}{Top-1}
& 0.6 & 79.2 & 10.9 & 42.2 & 69.9 & 0.9 & 38.4 & 58.8 & 1.3 & 27.9 & 48.0 & 8.5 & 13.2 & 34.0 & 5.0 & 4.6 & 65.2 & 0.4 & 41.0 \\
& 0.7 & 79.5 & 6.9 & 40.8 & 70.4 & 0.5 & 38.7 & 58.8 & 1.3 & 27.9 & 47.0 & 3.1 & 11.5 & 34.0 & 0.6 & 4.4 & 65.2 & 0 & 40.8 \\
& 0.8 & 80.5 & 2.6 & 40.0 & 70.4 & 0 & 38.5 & 58.8 & 0 & 27.6 & 46.6 & 1.0 & 10.9 & 34.0 & 0.6 & 4.4 & 65.2 & 0 & 40.8 \\
\cdashline{1-20}
\multirow{3}{*}{Entropy}
& 1.0 & 80.2 & 2.6 & 39.6 & 70.4 & 0 & 38.5 & 58.8 & 0 & 27.6 & 46.6 & 1.0 & 10.9 & 34.0 & 0.6 & 4.4 & 65.2 & 0 & 40.8 \\
& 1.2 & 79.9 & 4.6 & 40.1 & 70.4 & 0.5 & 38.7 & 58.8 & 1.3 & 27.9 & 46.6 & 1.2 & 10.9 & 34.0 & 0.6 & 4.4 & 65.2 & 0 & 40.8 \\
& 1.4 & 79.9 & 5.6 & 40.5 & 70.4 & 0.5 & 38.7 & 58.8 & 1.3 & 27.9 & 47.2 & 2.1 & 11.6 & 34.0 & 0.6 & 4.4 & 65.2 & 0 & 40.8 \\
\cdashline{1-20}
Explicit
& - & 41.6 & 0 & 0 & 31.9 & 0 & 0 & 31.3 & 0 & 0.1 & 35.7 & 1.7 & -0.2 & 29.6 & 1.3 & 0 & 24.4 & 0.4 & 0 \\
Dynathink 
& - & 50.3 & 1.1 & 8.8 & 61.3 & 2.9 & 30.2 & 55.3 & 7.6 & 26.1 & 45.1 & 9.1 & 10.1 & 34.0 & 3.1 & 4.5 & 50.1 & 2.2 & 26.3 \\
\textbf{CP-Router}
& - & 78.2 & 14.2 & \textcolor{red}{\textbf{42.7}} & 68.5 & 7.9 & \textcolor{red}{\textbf{39.7}} & 60.0 & 8.8 & \textcolor{red}{\textbf{31.6}} & 48.2 & 10.0 & \textcolor{red}{\textbf{13.7}} & 35.2 & 5.7 & \textcolor{red}{\textbf{5.9}} & 63.3 & 6.4 & \textcolor{red}{\textbf{41.6}} \\
\hline
\multicolumn{20}{c}{\textbf{Qwen-2.5-14B} and \textbf{DeepSeek-R1-Distill-Qwen-2.5-14B}} \\
LLM
& - & 46.9 & - & - & 31.5 & - & - &  36.2 & - & - & 44.9 & - & - & 25.2 & - & - & 30.4 & - & - \\
LRM
& - & 93.4 & - & 46.5 & 84.3 & - & 52.8 & 83.8 & - & 47.6 & 59.9 & - & 15.0 & 44.0 & - & 18.8 & 79.0 & - & 48.6 \\
\cdashline{1-20}
\multirow{4}{*}{Random}
& 0.2 & 84.4 & 19.8 & 46.8 & 75.4 & 19.3 & 54.3 & 76.8 & 16.0 & 48.3 & 56.8 & 19.7 & 14.8 & 40.4 & 19.9 & 18.9 & 70.5 & 19.4 & 49.8 \\
& 0.3 & 79.9 & 29.4 & 46.8 & 70.9 & 29.4 & 55.9 & 71.8 & 25.3 & 47.6 & 55.7 & 29.1 & 15.2 & 39.0 & 29.4 & 19.6 & 65.9 & 28.9 & 50.0 \\
& 0.4 & 75.6 & 39.3 & 47.3 & 65.7 & 39.1 & 56.1 & 66.0 & 34.5 & 45.5 & 54.6 & 39.3 & 16.0 & 37.2 & 38.7 & 19.6 & 61.3 & 38.9 & 50.5 \\
& 0.5 & 70.5 & 49.4 & 46.6 & 60.7 & 49.0 & 57.1 & 62.8 & 45.0 & 48.3 & 53.7 & 48.9 & 17.3 & 36.2 & 45.4 & 20.1 & 57.0 & 46.7 & 49.9 \\
\cdashline{1-20}
\multirow{3}{*}{Top-1}
& 0.6 & 89.8 & 23.3 & 55.9 & 85.2 & 5.8 & 57.0 & 81.3 & 8.8 & 49.4 & 60.1 & 16.5 & 18.2 & 45.2 & 3.4 & 20.7 & 79.4 & 0.6 & 49.3 \\
& 0.7 & 92.1 & 20.8 & 57.0 & 85.7 & 1.9 & 55.2 & 83.8 & 3.8 & 49.4 & 59.9 & 14.8 & 17.6 & 44.0 & 1.3 & 19.1 & 79.4 & 0.2 & 49.1 \\
& 0.8 & 93.4 & 11.6 & 52.6 & 86.6 & 0.9 & 55.6 & 83.8 & 2.5 & 48.8 & 59.9 & 6.0 & 15.9 & 44.0 & 0 & 18.8 & 79.4 & 0 & 49.0 \\
\cdashline{1-20}
\multirow{3}{*}{Entropy}
& 1.0 & 93.7 & 11.6 & 53.0 & 86.6 & 0.9 & 55.6 & 83.8 & 2.5 & 48.8 & 59.7 & 5.2 & 15.6 & 44.0 & 0 & 18.8 & 79.4 & 0 & 49.0 \\
& 1.2 & 92.7 & 16.8 & 55.1 & 85.7 & 1.9 & 55.2 & 83.8 & 3.8 & 49.4 & 59.9 & 10.8 & 16.8 & 44.0 & 0 & 18.8 & 79.4 & 0 & 49.0 \\
& 1.4 & 91.8 & 19.1 & 55.4 & 85.7 & 3.2 & 42.8 & 82.5 & 6.3 & 49.4 & 59.7 & 18.7 & 18.2 & 44.7 & 3.1 & 20.1 & 79.4 & 0.2 & 49.1 \\
\cdashline{1-20}
Explicit
& - & 67.0 & 45.5 & 36.9 & 57.4 & 44.0 & 46.3 & 57.5 & 41.3 & 36.3 & 51.8 & 38.4 & 11.2 & 38.1 & 38.4 & 21.0 & 58.3 & 46.0 & \textcolor{red}{\textbf{51.7}} \\
Dynathink
& - & 77.3 & 1.3 & 30.8 & 71.3 & 5.3 & 42.0 & 72.5 & 5.8 & 38.5 & 52.3 & 17.0 & 8.0 & 36.2 & 12.6 & 12.6 & 62.1 & 1.0 & 32.0\\
\textbf{CP-Router}
& - & 92.4 & 22.8 & \textcolor{red}{\textbf{58.9}} & 84.7 & 7.3 & \textcolor{red}{\textbf{57.4}} & 82.5 & 7.5 & \textcolor{red}{\textbf{50.1}} & 59.5 & 21.7 & \textcolor{red}{\textbf{18.7}} & 44.7 & 7.9 & \textcolor{red}{\textbf{21.2}} & 77.3 & 7.5 & 50.7 \\
\bottomrule
\end{tabular}
}
\caption{Main results for the Llama pairing (Llama-3.1-8B and DeepSeek-R1-Distill-Llama-8B) and the Qwen pairing (Qwen-2.5-14B and DeepSeek-R1-Distill-Qwen-14B) across various benchmarks. CP-Router achieves the highest token utility across all evaluated benchmarks with the Llama pairing, and on 5 out of 6 benchmarks with the Qwen pairing. Even in the single case where the Explicit baseline surpasses CP-Router in token utility, its accuracy is 19\% lower than that of CP-Router.}
\vspace{-4pt}
\label{tab:merged_main}
\end{table*}

To address the sensitivity to the choice of error rate $\alpha$, we propose an adaptive calibration strategy based on a novel criterion: Full and Binary Entropy (FBE). FBE is designed to capture the distribution of prediction set sizes from two complementary perspectives. The full entropy term measures the overall diversity across all prediction set sizes, helping to ensure well-separated uncertainty estimates. The binary entropy term captures the balance between singleton and non-singleton prediction sets, which is critical for avoiding routing load imbalance. Formally, FBE is defined as:
\begin{align}
\small
\text{FBE} &= \beta * H_{\text{full}} + H_{\text{binary}} \notag \\
&= -(\beta *\sum_{i} p_i \log p_i  \notag \\
& + \bigl[ p_{\text{s}=1} \log p_{\text{s}=1} + p_{\text{s} \neq 1} \log p_{\text{s} \neq 1} \bigr]),
\end{align}
\noindent
where \( p_i \) denotes the normalized frequency of prediction set size \( i \), 
\( p_{\text{s}=1} \) is the total frequency of singleton prediction sets (size = 1), and 
\( p_{\text{s} \neq 1} \) is the combined frequency of all non-singleton sets. 
\( H_{\text{full}} \) promotes diversity in uncertainty, while 
\( H_{\text{binary}} \) encourages a meaningful split between certain and uncertain cases. In our implementation, we perform a grid search over candidate values of $\alpha$ and compute FBE based on the prediction sets generated on the test set. We then select the $\alpha$ that maximizes FBE, as shown below:
\begin{align}
\small
\alpha^{*}=\mathop{\mathrm{argmax}}_{\alpha\in\mathcal{A}}\mathrm{FBE}.
\end{align}
\noindent With the calibrated error rate $\alpha^*$ selected via FBE, we can finalize the routing procedure: For each test prompt $x^{(j)}$, we use the LLM $f_\theta$ to compute a softmax distribution over the answer choices and construct the prediction set $C^{(j)}$. We then use the prediction set size-aware routing rule discussed in Section~\ref{vanillacp}. The full pseudocode of CP-Router is provided in Appendix~\ref{pseudocode}.

\section{Experiments}
\subsection{Experimental Setup}
\paragraph*{\textbf{Backbone LLMs and LRMs.}}
We use both small- and large-scale models, categorized into LLMs and LRMs. For LLMs, we include Llama-3-8B~\cite{grattafiori2024Llama}, Qwen-2.5-14B~\cite{yang2024qwen2}, and DeepSeek-V3~\cite{liu2024deepseek}. For LRMs, we use DeepSeek-R1-Distilled-Llama-3-8B, DeepSeek-R1-Distilled-Qwen-2.5-14B, and DeepSeek-R1~\cite{guo2025deepseek}.
\paragraph*{\textbf{Benchmarks.}}
We evaluate our method on seven MCQA benchmarks and a QA benchmark spanning several domains. For mathematics, we include MMLU-STEM~\cite{hendrycks2020measuring}, which covers elementary mathematics, high school statistics, and college mathematics; STEM-MCQA~\cite{TheWordsmiths2024}; and GSM8K~\cite{cobbe2021training}.
For logical reasoning, we use GPQA~\cite{rein2024gpqa} and LogiQA~\cite{liu2021logiqa}.
To evaluate multilingual robustness, we include CN-Chemistry~\cite{zhang2024chemllm}, a Chinese-language chemistry benchmark.
\paragraph*{\textbf{Baselines.}}  
Baselines fall into two categories:
\begin{itemize}[topsep=0pt, before=\setlength{\itemsep}{0pt}, parsep=0pt, leftmargin=*]
    \item \textbf{Single Model:} Using only an LLM or an LRM.
    \item \textbf{Hybrid Models with different routing strategies:} Including random routing, top-1 probability routing, response entropy routing, Dynathink (majority voting based routing)~\cite{pan-etal-2024-dynathink} and Explicit (explicit self-awareness routing). A detailed description of these baselines can be found in Appendix~\ref{baselines}.
\end{itemize}

\paragraph*{\textbf{Metrics.}}  
We evaluate the balance between effectiveness and efficiency using three metrics: (1) \textit{Acc} (Accuracy), which measures the correctness of final predictions; (2) \textit{TRR} (Token Reduction Ratio), defined as the proportion of output tokens saved compared to LRM, including those used for routing and inference/reasoning; and (3) \textit{$U_{\text{token}}$} (Token Utility), which captures the improvement in prediction accuracy per unit of token usage, defined in Equation~\ref{utoken} as the ratio of accuracy gain over token usage ratio of a given method relative to LRM. \textit{$U_{\text{token}}$} is introduced to facilitate a fairer comparison across methods with varying levels of token consumption.

\begin{equation}
\small
\label{utoken}
U_{\text{token}} = \frac{\textit{Acc - $Acc_{LLM}$}}{\textit{1 - TRR}}.
\end{equation}

\subsection{Main Results}
The performance of CP-Router on seven MCQA benchmarks, using  Qwen pairing and Llama  pairing, is summarized in Table~\ref{tab:merged_main}. Our method achieves the highest token utility on all 6 benchmarks with Llama pairing, and on 5 out of 6 with Qwen pairing, while maintaining competitive accuracy compared to all baselines. For example, on College Math with Llama pairing, we surpass LRM's accuracy by 1.2\% while also reducing token consumption. On the STEM-MCQA benchmark with Qwen pairing, although the Explicit baseline attains the highest token utility, its accuracy is significantly lower than that of CP-Router, with a 19\% decrease. This indicates that our method not only adapts to reduce token usage but also maintains—or even improves—the accuracy of the LRM. We also observe that different benchmarks correspond to different TRR values. For example, on Elementary Math, the Qwen pairing achieves a higher TRR of 22.8\%. In contrast, on High School Math and College Math—which are intuitively more difficult and require greater reasoning—lower TRR values are observed. This demonstrates that CP-Router can dynamically adjust its routing based on prediction uncertainty, making it adaptive rather than fixed.

\section{Analysis and Ablations}

\subsection{Routing Performance with Larger LLMs}
\label{strongllms}
\begin{figure}[t]
\centering
\includegraphics[width=1\linewidth]{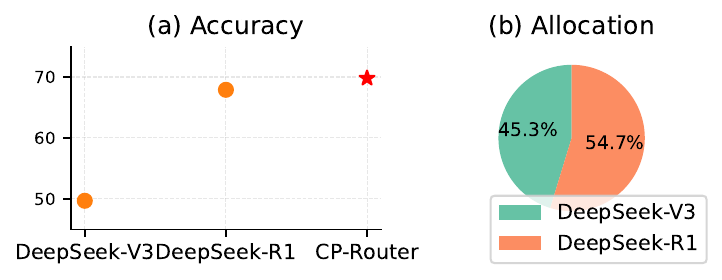}
    
\caption{Accuracy and prompt allocation of CP-Router on the GPQA benchmark using \textbf{DeepSeek-V3} and \textbf{DeepSeek-R1}. CP-Router improves overall accuracy while reducing token consumption by avoiding routing a portion of the prompts to the more expensive R1 model.}

    \label{deepseek}
    \vspace{-3mm}
\end{figure}

Due to computational constraints, experiments involving larger LLMs such as DeepSeek-V3 and DeepSeek-R1 are conducted only on the GPQA benchmark. The results in Figure~\ref{deepseek} show that the CP-Router outperforms both the standalone V3 and R1 models in terms of accuracy. Furthermore, it substantially reduces token consumption by routing only 54.7\% of the prompts to R1. These results demonstrate that the proposed method is effective not only for 7B and 14B models, but also scales well to larger, more robust ones, underscoring its model-agnostic nature.

\begin{table}[b]
    \centering
    \small

    \resizebox{0.45\textwidth}{!}{%
    \setlength{\tabcolsep}{5pt}
    \renewcommand{\arraystretch}{1.2}
    \begin{tabular}{lcccc}
        \toprule
        \multirow{2}{*}{\textbf{Pairings}} & 
        \textbf{Standard} & 
        \textbf{Reasoning} & 
        \multicolumn{2}{c}{\textbf{CP-Router}} \\
        & \textbf{Acc} & \textbf{Acc} & \textbf{Acc} & \textbf{TRR} \\
        \midrule
        Llama-8B & 45.3 & 46.0 & 47.6 & 18.9 \\
        Qwen-14B  & 82.1 & 71.6 & 82.9 & 68.0 \\
        \bottomrule
    \end{tabular}%
    }
\caption{Performance on CN-Chemistry. The LLM achieves higher accuracy in the 14B setting, while the LRM performs better in the 8B setting. In both cases, CP-Router further improves upon the stronger model’s accuracy.}
\label{tab:model_arc}
    \vspace{-6pt}
\end{table}

\subsection{Routing Performance when LLM is Stronger than LRM}
As shown in Table~\ref{tab:model_arc}, the CN-Chemistry benchmark presents a case where the LLM achieves higher accuracy in the 14B setting. We evaluate CP-Router under this condition and observe that it outperforms the LLM by more than 0.8\% in terms of accuracy. In the 8B pairing, the LRM achieves higher standalone performance, yet CP-Router routes effectively and still improves overall accuracy by 1.6\%. These results demonstrate that CP-Router is flexible and robust—it can adaptively leverage the strengths of the available models and even surpass the stronger LLM in the pairing. Furthermore, since CN-Chemistry is a Chinese-language benchmark, this further highlights the multilingual robustness of CP-Router.

\subsection{Ablations on FBE Components}
To assess the contribution of FBE in CP-Router, we conduct a comprehensive ablation study on its two components: full entropy and binary entropy. As shown in Table~\ref{ablations}, removing either component degrades routing performance in distinct ways. Excluding full entropy achieves substantial token savings (over 50\% TRR), but at the cost of a notable accuracy drop—for instance, from 68.5\% to 51.9\% on the High School Math subset. In contrast, using only binary entropy preserves accuracy (65.0\% on Elementary Math) but results in minimal token reduction (as low as 0.9\% TRR). Including both components strikes a favorable balance between efficiency and accuracy. Our results demonstrate that different entropy-based methods (lead to varying error rates), which affect prediction set sizes and, consequently, overall performance. This highlights the importance of reasonable automatic calibration. Based on these findings, we adopt a 3:1 weighting between full and binary entropy in all experiments.

\begin{table}[t]
    \centering
    \small

    \resizebox{0.47\textwidth}{!}{%
    \setlength{\tabcolsep}{5pt}
    \renewcommand{\arraystretch}{1.2} 
    \begin{tabular}{lcccccc}
        \toprule
        \multicolumn{1}{c}{\textbf{Full}} & 
        \multicolumn{1}{c}{\textbf{Binary}} & 
        \multicolumn{2}{c}{\textbf{Elementary Math}} & &
        \multicolumn{2}{c}{\textbf{High School Math}} \\
         & & \textbf{Acc} & \textbf{TRR} & &\textbf{Acc} & \textbf{TRR} \\
        \midrule

        \ding{55} & \ding{51} & 65.0 & 50.8 && 51.9 & 52.7 \\
        \ding{51} & \ding{55} & 78.2 & 10.6 && 69.0 & 0.9 \\
        1 & 1 & 77.9 & 14.5 && 58.3 & 33.4 \\
        2 & 1 & 78.2 & 14.2 && 66.2 & 13.4 \\
        3 & 1 & 78.2 & 14.2 && 68.5 & 7.9 \\
        \bottomrule
    \end{tabular}
    }
\caption{Ablation study results on the STEM-MMLU benchmark subsets Elementary Math and High School Math. Different entropy-based methods result in varying error rates, which affect prediction set sizes and ultimately influence overall performance.}
\vspace{-4pt} 
\label{ablations}
\end{table}

\subsection{Extension to General QA Format}
\begin{figure}[t]
\centering
    \includegraphics[width=1\linewidth]{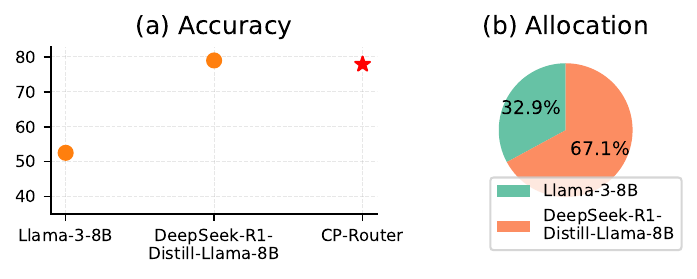}
    
\caption{Accuracy and prompt allocation of CP-Router on the open-ended QA benchmark GSM8K, using \textbf{Llama-3-8B} and \textbf{DeepSeek-R1-Distill-Llama-8B}, demonstrating its generalization to broader QA formats.}

    \label{gsm8k}
    \vspace{-2mm}
\end{figure}

We further extend our framework to the QA benchmark GSM8K, using the Llama-3-8B and DeepSeek-R1-Distill-Llama-8B pairing for evaluation. To adapt this open-ended format to our routing setup, we first prompt the LLM to generate its four most likely answers and then append an additional option, 'Others', to form a five-choice MCQ. We then apply the CP-Router procedure in the same manner as for the MCQA setting. As shown in Figure~\ref{gsm8k}, CP-Router achieves a competitive accuracy of 77.9\%, trailing the LRM by only 1.1\%, while saving reasoning-token consumption for 32.9\% of the questions. These results demonstrate the strong generalization capability of CP-Router to broader QA formats.

\subsection{CP-Router Generalizes Across Diverse Model Pairings}
\begin{figure}[h]
\centering
    \includegraphics[width=\linewidth]{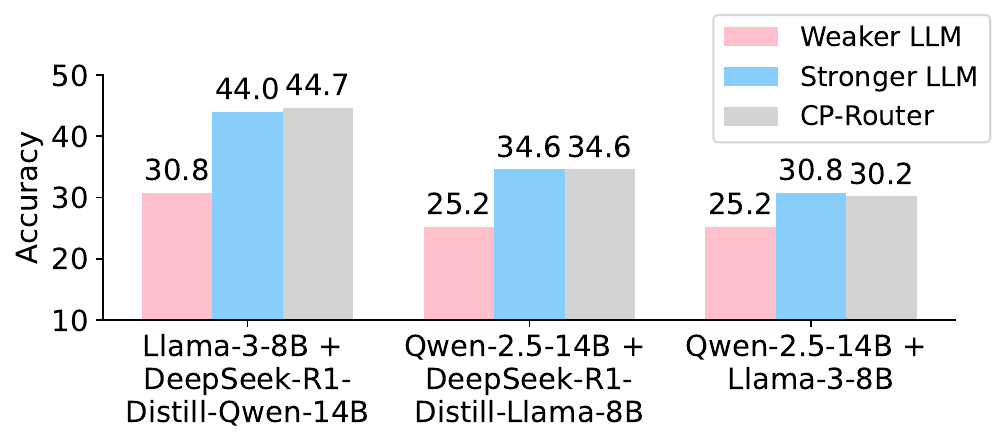}
    
\caption{CP-Router performance on GPQA under different model pairings. We evaluate both cross-family (e.g., Qwen LLM–Llama LRM) and intra-capability (e.g., Llama LLM-Qwen LLM) pairings to validate the model-agnostic nature of CP-Router. Results show CP-Router generalizes effectively across diverse pairings.}

    \label{bind}
    \vspace{-3mm}
\end{figure}

CP-Router is designed to be model-agnostic, capable of routing prompts across arbitrary model pairings. To evaluate its generalization ability, we conduct experiments on GPQA using both cross-family pairings (e.g., Qwen LLM–Llama LRM) and intra-capability pairings (e.g., Qwen LLM-Llama LLM), as shown in Figure~\ref{bind}. Across all configurations, CP-Router demonstrates robust performance. For instance, in the Llama-3-8B and Distill-Qwen-14B pairing, CP-Router even slightly surpassing LRM. In the Qwen-2.5-14B and Distill-Llama-8B pairing, it matches the LRM’s accuracy (34.6\%) while substantially outperforming LLM (25.2\%). Even in the Qwen-2.5-14B and Llama-3-8B pairing—where both models are LLMs—CP-Router remains competitive. These results show that CP-Router generalizes effectively across both architectural and capability dimensions.

\section{Related Work}
\subsection{LLM Routing}
LLM routing has received attention but remains underdeveloped. Training-based methods such as \citet{ong2024routellm} and \citet{aytes2025sketch} rely on classifiers to dispatch prompts, while \citet{chuang2024learning} introduce uncertainty-specialized tokens to enable confidence-aware routing. These methods require substantial training, limiting generalizability across diverse pairings. Beyond learned classifiers, \citet{pan-etal-2024-dynathink} use majority voting to route prompts with consistent outputs to lightweight paths, though repeated sampling is computationally costly. Recently, models like Claude 3.7 Sonnet~\citep{anthropic_37_2025} and Qwen3~\citep{qwen_3_2025} offer manual control over reasoning modes, shifting the routing burden to users who may lack the expertise to make effective decisions.

\subsection{Conformal Prediction in LLMs}
CP~\citep{vovk2005algorithmic, zhou2024conformal} has been applied to a range of natural language understanding tasks~\citep{dey2022conformal, maltoudoglou2020bert, pmlr-v204-ernez23a} prior to the rise of LLMs. More recent efforts have extended CP to LLMs, aiming to provide statistically rigorous uncertainty estimates. \citet{kumar2023conformal} and \citet{quach2023conformal} are among the first to adapt CP for LLMs. \citet{su-etal-2024-api} further addresses the limitations of API-based LLMs by introducing sampling-based nonconformity scores. \citet{wang2024conu} proposes ConU to incorporate self-consistency into CP to ensure correctness coverage for LLMs. \citet{cherian2024large} develops adaptive conformal inference techniques that offer conditional validity guarantees. While they improve CP’s reliability for LLMs, its application in routing remains unexplored.

\section{Conclusion}
This paper introduces CP-Router, a training-free and model-agnostic routing mechanism grounded in CP for efficient prompt routing between LLM and LRM. CP-Router dynamically selects either an LLM or an LRM based on the size of CP prediction sets, with error rates calibrated via FBE to more effectively differentiate uncertainty across inputs. Extensive experiments across diverse tasks demonstrate that CP-Router substantially reduces token usage while maintaining—and occasionally exceeding—the accuracy of LRMs. We further show that CP-Router generalizes across model scales and families, and can be extended beyond the MCQA task. We hope our findings inspire future research in uncertainty-aware LLM routing.


\bibliography{custom}


\appendix

\newpage
\section{Pseudocode}
\label{pseudocode}
We provide the pseudocode of CP-Router in Method~\ref{pseudocode-routing}.
\begin{method}[h]

\small

\caption{CP-Router}\label{pseudocode-routing}
\begin{algorithmic}[1]
\Require Prompt $x^{(i)}$, Ground truth $y^{(i)}$, LLM $f_{\theta}$, LRM $f_{\phi}$, Answer options $\mathcal{A}$, Calibration set $\mathcal{D}_{\text{cal}}$, Test set $\mathcal{D}_{\text{test}}$, Threshold $\tau$, Candidate error rates $\{\alpha_k\}$

\State {\textbf{Step 1: Calibration (Compute nonconformity scores)}}
    \For{$(x^{(i)}, y^{(i)}) \in \mathcal{D}_{\text{cal}}$}
        \State Get logits $z^{(i)} = f_{\theta}(x^{(i)})$
        \State Compute softmax: $p^{(i)} = \text{Softmax}(z^{(i)})$
        \State Compute nonconformity score: $s^{(i)} = 1 - p^{(i)}[y^{(i)}]$
    \EndFor

\State {\textbf{Step 2: Select best $\alpha$ by maximizing FBE on test set}}
    \For{each candidate $\alpha_k$}
        \State Compute $\hat{q}_{\alpha_k}$ = Quantile$\left(\{s^{(i)}\}, 1 - \alpha_k\right)$
        \For{$x^{(j)} \in \mathcal{D}_{\text{test}}$}
            \State Get logits $z^{(j)} = f_{\theta}(x^{(j)})$, softmax $p^{(j)}$
            \State Prediction set: $S^{(j)}_{\alpha_k} = \{a \in \mathcal{A} : 1 - p^{(j)}[a] \leq \hat{q}_{\alpha_k}\}$
        \EndFor
        \State Compute prediction set size distribution $\{p_i\}$ across test set
        \State $H_{\text{full}} = - \sum_i p_i \log p_i$  
        \State $H_{\text{binary}} = - (p_{s=1} \log p_{s=1} + p_{s\neq1} \log p_{s\neq1})$
        \State $\text{FBE}_{\alpha_k} = \beta \cdot H_{\text{full}} + H_{\text{binary}}$
    \EndFor
    \State Choose $\alpha^* = \arg\max_{\alpha_k} \text{FBE}_{\alpha_k}$
    \State Compute final $\hat{q}_{\alpha^*}$ from $\{s^{(i)}\}$

\State {\textbf{Step 3: Routing based on selected $\alpha^*$}}
    \For{$x^{(j)} \in \mathcal{D}_{\text{test}}$}
        \State $p^{(j)} = \text{Softmax}(f_{\theta}(x^{(j)}))$
        \State $S^{(j)} = \{a \in \mathcal{A} : 1 - p^{(j)}[a] \leq \hat{q}_{\alpha^*}\}$
        \If{$|S^{(j)}| \leq \tau$}
            \State \textbf{Route to LLM:} $\hat{y}^{(j)} = \arg\max_{a \in \mathcal{A}} p^{(j)}[a]$
        \Else
            \State \textbf{Route to LRM:} $\hat{y}^{(j)} = f_{\phi}(x^{(j)})$
        \EndIf
    \EndFor

\end{algorithmic}
\end{method}

\section{Baselines}
\label{baselines}
We describe the baselines used in our experiments, which fall into two categories:

\begin{itemize}[topsep=0pt, before=\setlength{\itemsep}{0pt}, parsep=0pt, leftmargin=*]
\item \textbf{Single Model:} Either an LLM or an LRM is used independently, with greedy decoding employed to generate the final answer.
\item \textbf{Hybrid Models with Different Routing Strategies:}
\begin{enumerate}[topsep=0pt, itemsep=0pt, leftmargin=*]
    \item \textbf{Random Routing:} Prompts are randomly routed to either the LLM or LRM. A predefined threshold controls the proportion directed to the LRM.
    
    \item \textbf{Top-1 Probability Routing}: The model selects the option with the highest probability as a proxy for confidence. If this probability exceeds a predefined threshold, the LLM is used; otherwise, the LRM is chosen.
    
    \item \textbf{Response Entropy Routing:} The entropy of the option probabilities is computed. If the entropy exceeds a predefined threshold, the LRM is used; otherwise, the LLM is selected.
    
    \item \textbf{Dynathink}~\cite{pan-etal-2024-dynathink}: Based on repeated sampling, if more than half of the sampled answers agree on an option, the prompt is routed to the LLM; otherwise, it is routed to the LRM.
    
    \item \textbf{Explicit Self-Awareness Routing:} The LLM generates both the answer and a self-assessment of whether it requires deliberation by the LRM. If so, the LRM is invoked.
\end{enumerate}
\end{itemize}

\section{Case Study}
As shown in Tables~\ref{case1} and \ref{case2}, CP-Router correctly identifies prompts where the LLM is both confident (i.e., low uncertainty with a prediction set size of 1) and accurate, thereby avoiding unnecessary routing to the LRM, whose reasoning is lengthy, incurs high token costs, and ultimately yields incorrect answers. It also detects high-uncertainty prompts that the LLM answers incorrectly, and instead routes them to the LRM, which successfully produces the correct answers. These case studies qualitatively demonstrate the effectiveness of our method.
\begin{table*}[t]

\centering
\small
\renewcommand{\arraystretch}{1.2}
\begin{tabular}{>{\raggedright\arraybackslash}p{0.25\textwidth} >{\raggedright\arraybackslash}p{0.7\textwidth}}
\toprule
Question & In the past, we had a lot of unrealistic high-profile in moral propaganda, so that a lot of the population said one thing and made one behind the other, and split personality. Through thinking about this phenomenon, some scholars have proposed that we should only ask ordinary people to abide by the "bottom line ethics". Based on your understanding, which of the following options is most appropriate as the definition of "bottom line ethics"? \\
\cdashlinelr{1-2}
Options & A. The bottom line ethics is not an ethics that requires selfless dedication. \newline
B. The bottom line ethics are some of the most basic and basic codes of conduct and rules that should be observed by ordinary people in a society. \newline
C. The bottom line ethics is not to steal or kill. \newline
D. If one compares human morality to a building, the bottom line ethics is the fundamental part of that building. \\
\cdashlinelr{1-2}
Llama-3-8B & "A": 0.044, "B": 0.853, "C": 0.038, "D": 0.065 \newline Prediction Set: {\color{red}[B]} \\
\cdashlinelr{1-2}
Routing? & \ding{55}. Use LLM. \\
\cdashlinelr{1-2}
DeepSeek-R1-Distill-Llama-8B (\ding{55} Not used, tokens saved) & Alright, so I'm trying to figure out what "bottom line ethics" means based on the given options. Let me start by understanding the context. The original question talks about past high-profile moral propaganda leading to people having split personalities, where they say one thing and do another. The scholars mentioned think that focusing on "bottom line ethics" might help.First, I need to break down each option and see which one best fits as a definition. Let's go through them one by one.Option A: "The bottom line ethics is not an ethics that requires selfless dedication." Hmm, this is saying that bottom line ethics don't require selflessness. But wait, isn't ethics usually about doing the right thing, which can include selfless acts? Maybe this is trying to say that it's a more basic level, not expecting much from people. Not sure yet.Option B: "The bottom line ethics are some of the most basic and fundamental codes of conduct and rules that should be observed by ordinary people in a society." This sounds like it's talking about the core, essential ethical principles that everyone should follow. That makes sense because if people can at least agree on these basics, society might function better. Option C: "The bottom line ethics is not to steal or kill." This seems a bit too specific. While not stealing or killing is definitely part of ethics, it's more like a subset rather than the entire bottom line. The bottom line should probably cover more than just these two actions. Option D: "If one compares human morality to a building, the bottom line ethics is the fundamental part of that building." This is using a metaphor, comparing morality to a building. The 'bottom line' in a building is the foundation, which is essential. So, applying that here, bottom line ethics would be the basic foundation upon which other ethical behaviors are built. This seems plausible.Now, comparing all options. Option B is about basic codes and rules, which aligns with the idea of a foundation. Option D uses a metaphor to describe the same concept, reinforcing the idea of a basic foundation. Option A talks about selflessness, which isn't directly part of the core ethics. Option C is too narrow. I'm torn between B and D. The question asks for the definition, and both B and D are good. However, the question mentions that the scholars proposed using "bottom line ethics," so the metaphor in D might be more specific and directly tied to the term. Also, the metaphor helps clarify what the bottom line is—it's the essential part without which the rest can't function. So, I think D is the best fit here. </think> The most appropriate definition of "bottom line ethics" is: <ans>D</ans>" \\
\cdashlinelr{1-2}
{\color{red}Correct Answer} & {\color{red}B} \\
\bottomrule
\end{tabular}
\caption{\label{case1} Case study A.}
\end{table*}

\begin{table*}[t]
\centering
\small
\renewcommand{\arraystretch}{1.2}
\begin{tabular}{>{\raggedright\arraybackslash}p{0.25\textwidth} >{\raggedright\arraybackslash}p{0.7\textwidth}}
\toprule
Question & Because the conclusion of quantum theory violates intuition, some scientists have different views on this theory. Although they try to strictly indicate that the assertion of quantum theory is inaccurate (that is, try to strictly falsify it), they find that the error lies in Usually within the acceptable statistical range. These results of quantum theory are different from those of competing theories, which makes it reasonable to accept quantum theory.Which of the following principles is most helpful to show the rationality of the above reasoning? \\
\cdashlinelr{1-2}
Options & If a scientific theory violates intuitive conclusions less than competing theories, then it should be accepted. \newline
B. A theory should not be considered reasonable before it is tried to strictly falsify. \newline
C. Only if the assertion of a theory has not been falsified by experiments can this theory be accepted. \newline
D. If you try to strictly falsify a theory, but the theory has withstood all the tests, then you should accept it. \\
\cdashlinelr{1-2}
Llama-3-8B & "A": 0.306, "B": 0.151, "C": 0.357, "D": 0.186 \newline Prediction Set: {\color{red}[A, C]} \\
\cdashlinelr{1-2}
Routing? & \ding{51}. Use LRM. \\
\cdashlinelr{1-2}
DeepSeek-R1-Distill-Llama-8B & 
Alright, so I'm trying to figure out this question about quantum theory and why it's accepted despite violating intuition. Let's break it down step by step.First, the question says that quantum theory's conclusion goes against our everyday intuition, but some scientists don't accept it. They try to strictly falsify it, meaning they look for experiments or evidence that would definitely show the theory is wrong. But when they do these tests, the results of quantum theory are still within what's considered an acceptable statistical range. This means the theory holds up even when rigorously tested. Because of this, it's reasonable to accept quantum theory over competing theories.Now, the question is asking which principle best shows the rationality of this reasoning. The options are A, B, C, D.Let's go through each option one by one.Option A: "If a scientific theory violates intuitive conclusions less than competing theories, then it should be accepted." Hmm. So, if a theory violates intuition less than others, accept it. But wait, quantum theory does violate intuition, so this might not apply directly. Also, just because it violates less doesn't necessarily mean it's correct. It depends on other factors like evidence and testing. So maybe not the best fit.Option B: "A theory should not be considered reasonable before it is tried to strictly falsify." This seems like it's saying you shouldn't accept a theory until you try to falsify it. But in the reasoning, the scientists are trying to falsify quantum theory and it's withstanding those tests. So, this principle is about the process—testing before acceptance. But does it directly show the rationality of accepting the theory? It's more about the process of evaluation.Option C: "Only if the assertion of a theory has not been falsified by experiments can this theory be accepted." This sounds like the principle of verificationism. You accept a theory only if experiments have not falsified it. In the scenario, quantum theory hasn't been falsified because it withstood tests, so according to this, it's acceptable. This seems to directly align with the reasoning. If a theory's predictions haven't been shown to be wrong, you can accept it tentatively.Option D: "If you try to strictly falsify a theory, but the theory has withstood all the tests, then you should accept it." This is similar to C but phrased differently. It's saying that if you try to falsify and fail, you accept the theory. In the case here, the theory hasn't been falsified, so you accept it. This also seems to fit the reasoning, but let me think about the wording.Wait, option C is a bit more general. It's saying that only if a theory hasn't been falsified can it be accepted. So, if experiments have falsified it, you don't accept it. But option D is more about the outcome of trying to falsify. It's a slightly different angle.In the given reasoning, the scientists tried to falsify quantum theory but found the error was within acceptable statistical range. So, they couldn't falsify it, hence they accept it. So, option D is about the outcome of falsification attempts leading to acceptance. Option C is more about the general condition—only accept if not falsified.The question is about showing the rationality of accepting quantum theory despite violating intuition. The reasoning relies on the fact that attempts to falsify it didn't work. So, option D directly addresses that outcome, making it the best fit.Wait, but earlier I thought C was also a contender. Let me clarify. The reasoning is: they tried to falsify, but the error was within the acceptable range, so they accept it. So, D is about the outcome—accepting because it withstood tests. C is more about the condition—only accept if not falsified.So, the reasoning is that because the theory hasn't been falsified, it's reasonable to accept it. Therefore, the best answer is D.</think>The reasoning supports accepting quantum theory because attempts to falsify it have not succeeded. This aligns with the principle that if a theory withholds falsification, it should be accepted. <ans>D</ans>\\
\cdashlinelr{1-2}
{\color{red}Correct Answer} & {\color{red}D} \\
\bottomrule
\end{tabular}
\caption{\label{case2} Case study B.}
\end{table*}

\end{document}